\documentclass{article}
\usepackage{spconf,amsmath,graphicx}
\usepackage{color}
\usepackage{multirow}
\usepackage{array}
\usepackage{xspace}
\usepackage{lineno,hyperref}
\usepackage{soul}
\usepackage{amsfonts,amssymb}
\usepackage{cleveref}

\let\OLDthebibliography\thebibliography
\renewcommand\thebibliography[1]{
  \OLDthebibliography{#1}
  \setlength{\parskip}{0pt}
  \setlength{\itemsep}{0pt plus 0.3ex}
}

\newcommand{\pink}[1]{\textcolor{magenta}{#1}}

\usepackage[normalem]{ulem}
\newcommand\system{SGENet\xspace}
\newcommand\etal{{\it et~al.}}

\begin{document}\sloppy

\def\x{{\mathbf x}}
\def\L{{\cal L}}

\title{Efficient Scene Text Image Super-resolution with Semantic Guidance}
%
\name{LeoWu TomyEnrique$^1$, Xiangcheng Du$^1$, Kangliang Liu$^1$, Han Yuan$^2$, Zhao Zhou$^{1,2}$, Cheng Jin$^{1,3*}$\thanks{$^*$Corresponding author.}\thanks{This work was supported by the National Archives Administration of China Research Program (2021-X-25).}}
\address{
    $^1$School of Computer Science, Fudan University, Shanghai, China\\
    $^2$Videt Technology, Shanghai, China\\
    $^3$Innovation Center of Calligraphy and Painting Creation Technology, MCT, China
}
%
%
%
%
\maketitle

\begin{abstract}
Scene text image super-resolution has significantly improved the accuracy of scene text recognition. However, many existing methods emphasize performance over efficiency and ignore the practical need for lightweight solutions in deployment scenarios.  Faced with the issues, our work proposes an efficient framework called~\system to facilitate deployment on resource-limited platforms.
~\system contains two branches: super-resolution branch and semantic guidance branch. We apply a lightweight pre-trained recognizer as a semantic extractor to enhance the understanding of text information. 
Meanwhile, we design the visual-semantic alignment module to achieve bidirectional alignment between image features and semantics, resulting in the generation of high-quality prior guidance. 
We conduct extensive experiments on benchmark dataset, and the proposed~\system achieves excellent performance with fewer computational costs.
Code is available at \pink{\href{https://github.com/SijieLiu518/SGENet}{https://github.com/SijieLiu518/SGENet}}.
\end{abstract}

\section{Introduction}
\label{sec:intro}

Scene Text Image Super-Resolution (STISR) aims to enhance the resolution and quality of text images captured from real-world scenes. Due to the proliferation of low-resolution text images captured under low lighting or motion blur, STISR has drawn a lot of attention. As preprocessing task for text recognition, STISR~\cite{wang2020scene,Chen2021SceneTT,ma2023text,ma2022text} technology is crucial for scene text recognition~\cite{Du2022SVTRST, xie2022toward, Fang2021ReadLH}, document analysis~\cite{zhong2019publaynet, biswas2021beyond}, and text extraction and analytics~\cite{guo2019eaten, wang2019efficient}.

\begin{figure}[t]
    \centering
    \includegraphics[width=.95\linewidth]{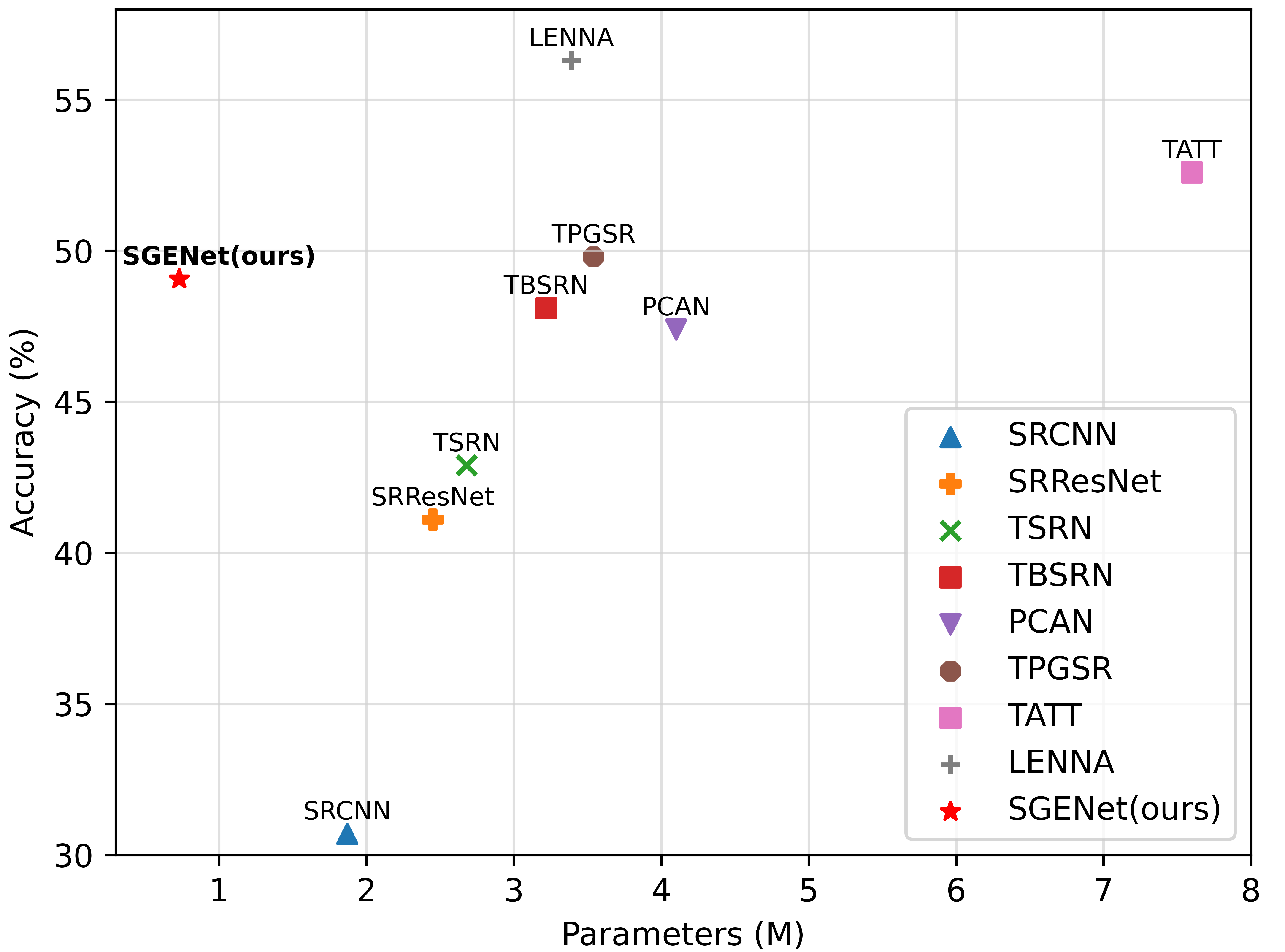}
    \caption{Total number of parameters $vs.$ scene text recognition accuracy in different STISR networks. Our model achieves satisfactory results considering that it has fewer parameters.}
    \label{fig:fig1}
\end{figure}

Recently, many methods have been proposed to enhance the scene text image quality which aims to improve the performance of STR. For example, Dong~\etal~\cite{dong2015boosting} utilize SRCNN~\cite{dong2015image} as the backbone for text images recovery. Tran~\etal~\cite{tran2019deep} introduce LapSRN~\cite{lai2017deep} to STISR task and enhance edges in super-resolution image. In these methods, STISR models treat scene text images as general object without taking text-specific characteristics. After that, some methods use text prior information to guide scene text image reconstruction. TPGSR~\cite{ma2023text} adopts the character probability sequence to reconstruct text characters more effectively. TATT~\cite{ma2022text} refines the text structure by imposing structural consistency between the recovered regular and deformed texts.
Although achieving excellent performance, the above methods have limitations. They rely on complicated networks and may require prior knowledge of extensive pre-trained models, which makes their application challenging on resource-limited devices.
Each STISR model consumes much time due to significant parameters and high computational complexity during the inference process. Therefore, it is crucial to design an efficient and accurate STISR model.

\begin{figure*}[t]
    \centering
    \includegraphics[width=.98\linewidth]{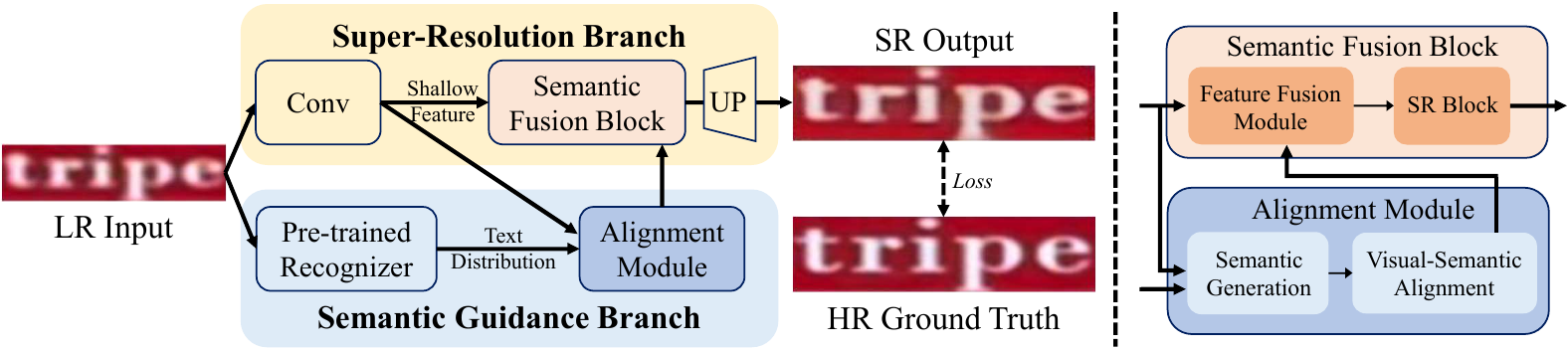}
    \caption{The overall architecture of~\system. It consists of two branches, the semantic guidance branch and the super-resolution branch. The output of the semantic guidance branch is used to guide super-resolution reconstruction.}
    \label{fig:pipline}
  \end{figure*}

In this paper, we propose an efficient STISR model (\system) that contains fewer parameters while achieving satisfying performance within low computational complexity as shown in Fig.~\ref{fig:fig1}.
The framework contains super-resolution branch and semantic guidance branch. Specifically, we first introduce a pre-trained text recognizer to generate text distribution in the semantic guidance branch.
Then we use the text distribution and image shallow features as the input of alignment module for semantic generation.
To achieve better alignment~\cite{Guo2023TowardsRS} between different modalities in a progressive manner, we perform bidirectional-cross attention layers on visual-semantic to generate high-level semantic prior.
Finally, the high-level semantic prior is concatenated with image features and then passed to the feature fusion module, guiding the super-resolution reconstruction process.
Our contributions can be summarized as follows:
\begin{itemize}
    \item We proposed an efficient scene text image super-resolution network that obtains an optimal balance between the efficiency and performance of the model.
    \item In the semantic guidance branch, we modeling global shallow feature and semantic information to enhance the comprehension of contextual information.
    \item Compared with existing multi-branch frameworks that necessitate a pre-trained recognizer, our framework offers simplicity, efficient training, and reduces computational resource requirements.
\end{itemize}

\section{Related Work}

\label{sec:related}
\vspace{0.03in}
\noindent\textbf{Scene Text Recognition.}
Scene text recognition has made great progress in recent years, and it aims to extract text content from the input images. Considering text recognition as an image-to-sequence problem, CRNN~\cite{Shi2015AnET} combines CNN and RNN to model sequential features of text images. It is trained with CTC~\cite{Graves2006ConnectionistTC} loss to align the predicted sequence and the target sequence. Recently, text recognition is enhanced by SRN~\cite{Yu2020TowardsAS} and ABINet~\cite{Fang2021ReadLH} through the incorporation of language models, while MATRN~\cite{Na2021MultimodalTR} leverages visual-semantic multi-modality to improve its performance. However, low-resolution text images still pose challenges for these methods. Therefore, it is necessary to use the STISR method to obtain easily recognizable scene text images.

\vspace{0.03in}
\noindent\textbf{Scene Text Image Super Resolution.}
STISR aims to enhance the visual quality and legibility specifically for scene text images. The earlier SISR methods can be readily employed in STISR, allowing for direct adoption of CNN architectures.
Dong~\etal~\cite{dong2015boosting} extended SRCNN~\cite{dong2015image} to text images, and resulting in remarkable performance. PlugNet~\cite{Mou2020PlugNetDA} employs a pluggable super-resolution unit to upsample text images within the feature domain. Recently, a real scene text SR dataset was proposed, termed TextZoom~\cite{wang2020scene}. It was more authentic and challenging than synthetic data. PCAN~\cite{Zhao2021SceneTI} improved the network structure through the use of parallel contextual attention. Chen~\etal~\cite{Chen2021SceneTT} enhanced the model to pay more attention to the position and the content of each character by the position-aware module and content-aware module.

LEMMA~\cite{Guo2023TowardsRS} use a large text recognizer (ABINet)~\cite{Fang2021ReadLH} to improve the STISR performance, and propose the multimodal alignment module to alignment the different modalities.
Different from LEMMA~\cite{Guo2023TowardsRS}, we utilize a lightweight recognizer to generate semantic guidance. Moreover, we provide semantic guidance based on global visual features. In order to further reduce the number of parameters of the network, we reduce redundant blocks in the super-resolution branch.

\section{Approach}
\label{sec:method}

\subsection{Overview}
\label{subsec:overall}
The overall pipeline of~\system as shown in Fig.~\ref{fig:pipline}.
Given a low-resolution image $I_{LR}\in\mathbb{R}^{H\times W \times 3}$, where $H$ and $W$ are the height and width. The task aims to produce super-resolution image $I_{SR}\in{\mathbb{R}^{sH\times sW \times 3}}$, where $s$ is the scale factor.
We first use CNN to extract shallow features $f_{s}$ from the input image $I_{LR}$. Specifically, the proposed pipeline consists of two branches, the super-resolution branch and the semantic guidance branch. As for the semantic guidance branch, the pre-trained text recognizer takes $I_{LR}$ as input to generate the text distribution. Then visual-semantic alignment module align text distribution and $f_s$.
As for the super-resolution branch, $N$ semantic fusion blocks receive $f_s$ and high-level guidance to reconstruct the high-resolution scene text image.

\subsection{Semantic Guidance Branch}
The branch initially uses the $I_{LR}$ to generate the text distribution which will be fed into the semantic generation module to generate semantic features $h_{t}$. The semantic feature $h_{t}$ and show visual feature $f_s$ are employed in the alignment module.

\vspace{0.03in}
\noindent\textbf{Semantic Generation Module.}
Following~\cite{Guo2023TowardsRS}, we first perform self-attention operation on the text distribution obtained from the pre-trained recognizer. The rich semantic features $h_{t}$ and shallow image feature $f_s$ are fed into following visual semantic alignment module to achieve alignment by cross attention layers.

\vspace{0.03in}
\noindent\textbf{Visual-Semantic Alignment Module.}
We employ cross-attention to achieve alignment between the visual and semantic information. Different from self-attention, cross-attention can process different modalities, such as text, images. Specifically, the visual-semantic alignment module contains a self-attention layer and two cross-attention layers.
For the first cross-attention layer, we use $h_{t}$ as query, $f_{s}$ as key and value to calculate the relationship between each character and different region in the image:
\begin{equation}
    \begin{aligned}
        h^{'}_{i} &= \texttt{LN}(\texttt{MultiHead}(h_{i-1}, f_{s}, f_{s}) + h_{i-1}) \\
        h_{i} &= \texttt{LN}(\texttt{MLP}(h^{'}_{i})+h^{'}_{i})
    \end{aligned}
\end{equation}
where $i$ denotes the $i$-th attention block, $h_0$ is $h_t$ and $h_{i-1}$ is the output of previous block.

We use the output of first cross-attention layer as the key ($h_{ca}$) of second cross-attention layer, $f_{s}$ as query and $h_{t}$ as value to allow each element of $f_{s}$ can find which text feature it should attend by using $h_{ca}$. Through the cross-attention layers, we can align the text information and visual feature and obtain high-level guidance $h_g$.

\subsection{Super-Resolution Branch}
The branch aims to reconstruct HR scene text image from the shallow features $f_s$ and high-level guidance. Following~\cite{Guo2023TowardsRS}, we also adopt feature fusion module to combines visual feature and high-level guidance. Image feature $f_s$ and high-level guidance $h_g$ are first concatenated by three parallel $1 \times 1$ convolution and generate different features $f^1_s$, $f^2_s$ and $f^3_s$. Then we perform the channel attention operation on $f^1_s$ to obtain the attention score and multiple the score with $f^2_s$ to generate channel attention feature, which will be added to $f^3_s$ to get the final result. We formulate the process as follows:
\begin{align}
    f = f^3_s + f^2_s \otimes \texttt{CA}(f^1_s)
\end{align}
where \texttt{CA} denotes the channel attention mechanism.

Then the fused feature $f$ is input to Sequential Recurrent Block (SRB)~\cite{wang2020scene} to build deeper and sequential dependencies. Finally, the super-resolution image is generated by pixel shuffling following~\cite{wang2020scene,Guo2023TowardsRS}.

\subsection{Training Loss}
During training phase, the following three losses are adopted:

\vspace{0.03in}
\noindent\textbf{Reconstruction loss.} We adopt MSE loss to perform image reconstruction supervision in the super-resolution branch:
\begin{equation}
    L_{rc} = \texttt{MSE}(I_{SR} - I_{HR})
\end{equation}

\vspace{0.03in}
\noindent\textbf{Recognition loss.} Inspired by~\cite{Chen2021SceneTT}, position-aware and the content-aware loss is used to supervise the learning of language knowledge:
\begin{equation}
    \begin{aligned}
        &L_{pos} = ||\texttt{A}_{HR}-\texttt{A}_{SR}||_{1} \\
        &L_{con} = \texttt{WCE}(\texttt{P}_{SR},y_{label}) \\
        &L_{re} = \lambda_{pos}L_{pos} + \lambda_{con}L_{con}
    \end{aligned}
\end{equation}
where $\texttt{A}$ denotes the attention map extracted from the middle layer of the Transformer, \texttt{P} denotes probability distribution. $\texttt{WCE}$ denotes weighted cross-entropy.

\vspace{0.03in}
\noindent\textbf{Fine-tuning loss.} Fine-tuning have shown that is better than fixed parameters~\cite{ma2022text, Zhao2022C3STISRST}. Therefore, we use the cross-entropy loss to adapt the text recognizer to low resolution inputs:
\begin{equation}
    L_{ft}=-\sum^{n}_{i=1}y_{label}\texttt{log}(\texttt{P}_{pr})
\end{equation}
where $\texttt{P}_{pr}$ denotes probability distribution predicted by the pre-trained recognizer.

The overall loss function is described as follows:
\begin{equation}
    L=L_{rc}+\alpha_{1}L_{re}+\alpha_{2}L_{ft}
\end{equation}
where $\alpha_1$ and $\alpha_2$ are hyperparameters.

\section{Experiments}
\label{sec:exp}

\begin{table*}[t]
  \center
  \caption{Performance comparison on the three subsets in TextZoom. The recognition accuracies are evaluated using the models released officially by ASTER \cite{Shi2019ASTERAA}, MORAN \cite{Luo2019AMR} and CRNN \cite{Shi2015AnET}.}
  \label{tab:performance}
    \resizebox{\textwidth}{!}{
    \begin{tabular}{c|cccc|cccc|cccc|cc}
      \hline \multirow{2}{*}{ Method } & \multicolumn{4}{c|}{ ASTER \cite{Shi2019ASTERAA} } & \multicolumn{4}{c|}{ MORAN \cite{Luo2019AMR} } & \multicolumn{4}{c|}{ CRNN \cite{Shi2015AnET} } &  \multirow{2}{*}{Params(M)} &  \multirow{2}{*}{FLOPs(G)}\\
      \cline { 2 - 13 } & Easy & Medium & Hard & Avg & Easy & Medium & Hard & Avg & Easy & Medium & Hard & Avg \\
      \hline Bicubic & $64.7 \%$ & $42.4 \%$ & $31.2 \%$ & $47.2 \%$ & $60.6 \%$ & $37.9 \%$ & $30.8 \%$ & $44.1 \%$ & $36.4 \%$ & $21.1 \%$ & $21.1 \%$ & $26.8 \%$ & - & - \\
      SRCNN \cite{dong2015image} & $69.4 \%$ & $43.4 \%$ & $32.2 \%$ & $49.5 \%$ & $63.2 \%$ & $39.0 \%$ & $30.2 \%$ & $45.3 \%$ & $38.7 \%$ & $21.6 \%$ & $20.9 \%$ & $27.7 \%$ & 1.87 & 0.13\\
      SRResNet \cite{Ledig2016PhotoRealisticSI} & $69.4 \%$ & $47.3 \%$ & $34.3 \%$ & $51.3 \%$ & $60.7 \%$ & $42.9 \%$ & $32.6 \%$ & $46.3 \%$ & $39.7 \%$ & $27.6 \%$ & $22.7 \%$ & $30.6 \%$ & 2.45 & 0.68 \\
      TSRN \cite{wang2020scene} & $75.1 \%$ & $56.3 \%$ & $40.1 \%$ & $58.3 \%$ & $70.1 \%$ & $53.3 \%$ & $37.9 \%$ & $54.8 \%$ & $52.5 \%$ & $38.2 \%$ & $31.4 \%$ & $41.4 \%$ & 2.68 & 0.91 \\
      TBSRN \cite{Chen2021SceneTT} & $75.7 \%$ & $59.9 \%$ & $41.6 \%$ & $60.0 \%$ & $74.1 \%$ & $57.0 \%$ & $40.8 \%$ & $58.4 \%$ & $59.6 \%$ & $47.1 \%$ & $35.3 \%$ & $48.1 \%$ & 3.22 & 1.22 \\
      LEMMA \cite{Guo2023TowardsRS} & $81.1 \%$ & $66.3 \%$ & $47.4 \%$ & $66.0 \%$ & $77.7 \%$ & $64.4 \%$ & $44.6 \%$ & $63.2 \%$ & $67.1 \%$ & $58.8 \%$ & $40.6 \%$ & $56.3 \%$ & 3.39 & 6.70 \\ 
      TPGSR \cite{ma2023text} & $77.0 \%$ & $60.9 \%$ & $42.4 \%$ & $60.9 \%$ & $72.2 \%$ & $57.8 \%$ & $41.3 \%$ & $57.8 \%$ & $61.0 \%$ & $49.9 \%$ & $36.7 \%$ & $49.8 \%$ & 3.54 & 1.01 \\
      PCAN \cite{Zhao2021SceneTI} & $77.5 \%$ & $60.7 \%$ & $43.1 \%$ & $61.5 \%$ & $73.7 \%$ & $57.6 \%$ & $41.0 \%$ & $58.5 \%$ & $59.6 \%$ & $45.4 \%$ & $34.8 \%$ & $47.4 \%$ & 4.10 & 1.85 \\
      TATT \cite{ma2022text} & $78.9 \%$ & $63.4 \%$ & $45.4 \%$ & $63.6 \%$ & $72.5 \%$ & $60.2 \%$ & $43.1 \%$ & $59.5 \%$ & $62.6 \%$ & $53.4 \%$ & $39.8 \%$ & $52.6 \%$ & 7.60 & 1.26 \\ \hline
      \system(ours) & $75.8 \%$ & $60.7 \%$ & $45.0 \%$ & $61.4 \%$ & $71.5 \%$ & $56.2 \%$ & $41.4 \%$ & $57.3 \%$ & $59.4\%$ & $47.9\%$ & $37.7\%$ & $49.0\%$ & 0.73 & 0.98 \\
      \hline
      \end{tabular}
    }
    \vspace{-0.1in}
  \end{table*}

\subsection{Experimental Setting}
\noindent\textbf{Dataset.} TextZoom~\cite{wang2020scene} are collected from two image super-resolution datasets, including RealSR~\cite{Cai2019TowardRS} and SR-RAW~\cite{Zhang2019ZoomTL}. TextZoom contains 17,367 LR-HR pairs for training and 4,373 pairs for testing. According to different focal lengths of digital cameras, the samples for testing can be divided into three subsets, namely easy (1,619 samples), medium (1,411 samples) and hard (1,343 samples). LR images are resized to 16 $\times$ 64 and HR images are resized to 32 $\times$ 128. 

\vspace{0.03in}
\noindent\textbf{Implementation Details.} Our \system is implemented in PyTorch. All experiments are conducted on four NVIDIA TITAN Xp GPUs with 12GB memory. We adopt Adam~\cite{Kingma2014AdamAM} optimizer to train the model with batch size 64 for 500 epochs. The learning rate is set to 0.0001 for the super-resolution. Moreover, we apply SVTR-T~\cite{Du2022SVTRST} as the pre-trained text recognizer, and the parameters are frozen.

\begin{figure}[t]
  \centering
  \includegraphics[width=.96\linewidth]{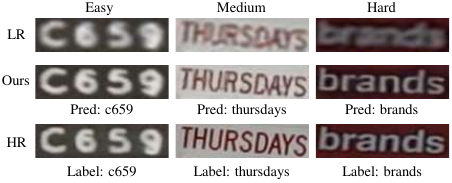}
  \caption{Visualization results of~\system on TextZoom dataset.}
  
  \label{fig:fig3}
\end{figure}

\subsection{Comparing with State-of-the-Arts Methods}
In Tabel~\ref{tab:performance}, we compare our model with other existing super-resolution models on three recognition models, including CRNN~\cite{Shi2015AnET}, ASTER~\cite{Shi2019ASTERAA}, and MORAN~\cite{Luo2019AMR}. Compared with the baseline TSRN, our model boosts average accuracy by 3.1\% on ASTER, 2.5\% on MORAN, and 7.6\% on CRNN. In contrast to the state-of-the-art methods, \system experiences only a minor decrease in performance, while striking an improved balance between efficiency and effectiveness.

Meanwhile, we evaluate \system and other STISR methods using model parameter and computational complexity. When compared with the SOTA solutions, \system exhibits only a marginal reduction in performance, which is significantly outweighed by the benefits it offers in terms of improved efficiency. Our experiments show that \system has the smallest number of parameters. Due to \system having low computational complexity, the inference process is efficient. Especially, \system reduces 85\% computational complexity compared with the state-of-the-art method (LEMMA), which makes it easily deployable on resource-limited devices. The balance between excellent performance and efficient computation makes \system a practical choice for scene text image super-resolution tasks. We visualize the scene text image super-resolution results in Fig.~\ref{fig:fig3}.

\subsection{Ablation Study}
We conduct ablation studies to assess \system by using CRNN recognizer~\cite{Shi2015AnET}.

\begin{table}[t]
  \centering
  \caption{Different number of SRBs.}
  \label{tab:srbs}
  \begin{tabular}{c|cccc}
    \hline \multirow{2}{*}{ SRBs } & \multicolumn{4}{c}{ Accuracy } \\
    \cline { 2 - 5 }  & Easy & Medium & Hard & Avg \\ \hline 
    2 & 59.4\% & 47.9\% & 37.7\% & 49.0\% \\
    4 & 60.3\% & 48.9\% & 36.0\% & 49.2\% \\
    6 & 60.1\% & 47.7\% & 36.1\% & 48.7\% \\
    \hline
  \end{tabular}
  \vspace{-0.1in}
\end{table}

\vspace{0.03in}
\noindent\textbf{Block Number.} We evaluate the impact of the number of SRBs. Table~\ref{tab:srbs} reports the accuracy achieved for different SRB quantities of 2, 4, and 6. Notably, as the number of SRBs increases, there is no substantial improvement in performance. In fact, the accuracy even decreases when using 6 SRBs. To balance parameters and performance, we choose number of SRB is 2.

\begin{table}
  \centering
  \small
  \caption{Different pre-trained text recognizers.}
  \label{tab:rec}
  \begin{tabular}{c|cccc|c}
    \hline \multirow{2}{*}{ Rec } & \multicolumn{4}{c|}{ Accuracy } & \multirow{2}{*}{FLOPs(G)} \\
    \cline { 2 - 5 } & Easy & Medium & Hard & Avg \\
    \hline SVTR-T & 59.4\% & 47.9\% & 37.7\% & 49.0\% & 0.98 \\
    ABINet & 62.6\% & 49.8\% & 37.9\% & 50.9\% & 6.11 \\
    \hline
  \end{tabular}
  \vspace{-0.1in}
\end{table}

\noindent\textbf{Pre-Trained Text Recognizer.} We conduct different recognizer to evaluate the effect of pre-trained recognizer. The results in Table~\ref{tab:rec} show that stronger recognizer can boost the performance of STSR model. Using ABINet~\cite{Fang2021ReadLH} for text distribution can boost performance, but the huge model complexity is not conducive to building an efficient model.
ABINet exhibits a computational complexity that is 85\% greater than that of SVTR-T. However, the recognition accuracy has not been significantly improved.

\section{Conclusions}
\label{sec:conclusion}
In this paper, we present a Semantic Guidance Efficient network (\system) to facilitate deployment on resource-limited platforms.
\system consists of two branches: the super-resolution branch and the semantic guidance branch. We leverage a a lightweight pre-trained recognizer for semantic extraction, enhancing textual comprehension in images. Additionally, a visual-semantic alignment module establishes coherence between image features and semantics, generating high-quality prior guidance.
Extensive experiments on the TextZoom dataset show that \system achieves a superior trade-off between performance and efficiency with reduced computational resources.

\bibliographystyle{IEEEbib}
\bibliography{total}

\end{document}